\documentclass[sigconf]{acmart}
\AtBeginDocument{%
  }

\setcopyright{acmlicensed}
\copyrightyear{2026}
\acmYear{2026}
\acmDOI{XXXXXXX.XXXXXXX}
\acmConference[Preprint]{Preprint Under Review}{2026}{}
\acmISBN{978-1-4503-XXXX-X/2018/06}

\usepackage{hyperref}
\usepackage{xcolor}         
\usepackage{graphicx}
\usepackage{multirow}
\usepackage{adjustbox} 
\usepackage{array}
\usepackage{graphicx}
\usepackage{url} 
\usepackage[table]{xcolor}
\definecolor{rowgray}{gray}{0.95}
\usepackage[inkscapelatex=false]{svg}
\svgpath{{./}} 

\usepackage{enumitem}
\usepackage{subfig}
\usepackage[most]{tcolorbox}

\begin{document}

\title{Quantum-Audit: Evaluating the Reasoning Limits of LLMs on Quantum Computing}

\author{Mohamed Afane}
\authornote{Corresponding authors: \{mafane, jchen504\}@fordham.edu}
\email{mafane@fordham.edu}
\affiliation{%
  \institution{Fordham University}
  \city{New York}
  \state{New York}
  \country{USA}
}

\author{Kayla Laufer}
\email{kal1@fordham.edu}
\affiliation{%
  \institution{Fordham University}
  \city{New York}
  \state{New York}
  \country{USA}
}

\author{Wenqi Wei}
\email{wenqiwei@fordham.edu}
\affiliation{%
  \institution{Fordham University}
  \city{New York}
  \state{New York}
  \country{USA}
}

\author{Ying Mao}
\email{ymao41@fordham.edu}

\affiliation{%
  \institution{Fordham University}
  \city{New York}
  \state{New York}
  \country{USA}
}

\author{Junaid Farooq}
\email{mjfarooq@umich.edu}

\affiliation{%
  \institution{University of Michigan-Dearborn}
  \city{Dearborn}
  \state{Michigan}
  \country{USA}
}

\author{Ying Wang}
\email{ywang6@stevens.edu}

\affiliation{%
  \institution{Stevens Institute of Technology}
  \city{Hoboken}
  \state{New Jersey}
  \country{USA}
}

\author{Juntao Chen}
\authornotemark[1]
\email{jchen504@fordham.edu}
\affiliation{%
  \institution{Fordham University}
  \city{New York}
  \state{New York}
  \country{USA}
}
\renewcommand{\shortauthors}{Afane et al.}

\begin{abstract}
Language models have become practical tools for quantum computing education and research, from summarizing technical papers to explaining theoretical concepts and answering questions about recent developments in the field. While existing benchmarks evaluate quantum code generation and circuit design, their understanding of quantum computing concepts has not been systematically measured. Quantum-Audit addresses this gap with 2,700 questions covering core quantum computing topics. We evaluate 26 models from leading organizations. Our benchmark comprises 1,000 expert-written questions, 1,000 questions extracted from research papers using LLMs and validated by experts, plus an additional 700 questions including 350 open-ended questions and 350 questions with false premises to test whether models can correct erroneous assumptions. Human participants scored between 23\% and 86\%, with experts averaging 74\%. Top-performing models exceeded the expert average, with Claude Opus 4.5 reaching 84\% accuracy, though top models showed an average 12-point accuracy drop on expert-written questions compared to LLM-generated ones. Performance declined further on advanced topics, dropping to 73\% on security questions. Additionally, models frequently accepted and reinforced false premises embedded in questions instead of identifying them, with accuracy below 66\% on these critical reasoning tasks.
\end{abstract}




\maketitle

\section{Introduction}

Quantum computing has progressed significantly from theoretical research to experimental implementations with practical applications. Current quantum systems have rapidly evolved through successive technological breakthroughs from operating with just a few qubits to recently surpassing the 1000-qubit barrier~\cite{abughanem2025ibm}, enabling exploration of quantum algorithms and protocols that were previously confined to theoretical analysis. This technical advancement drives progress in quantum simulation~\cite{king2025beyond,halimeh2025cold,puig2025variational}, optimization problems~\cite{quinton2025quantum,phillipson2024quantum}, and cryptographic applications~\cite{sahu2024state,ralegankar2021quantum,kalaivani2021enhanced}. Beyond traditional quantum applications such as quantum simulation and cryptography, recent research explores its potential in finance~\cite{innan2024financial,grossi2022mixed}, healthcare~\cite{ur2023quantum, flother2023state}, computer vision~\cite{li2020quantum,afane2025atp,alrikabi2022face}, and wireless communication~\cite{narottama2021quantum,narottama2023quantum}, among other promising real-world applications.

In parallel, Large Language Models (LLMs) have become sophisticated tools that address complex challenges across many disciplines. These AI systems now approach or exceed human expert performance in areas such as cybersecurity~\cite{tihanyi2024cybermetric, afane2024next}, medical diagnosis~\cite{subedi2025reliability}, and legal reasoning~\cite{guha2023legalbench,kant2025towards}. As these two fields continue to evolve, their intersection becomes increasingly important for scientific communication, education, and research productivity.
Despite significant advances in both domains, we face a critical knowledge gap in evaluating LLMs' understanding of specialized quantum concepts. While extensive benchmarking exists across numerous related domains, including mathematics~\cite{gao2024omni, fang2024mathodyssey}, physics \cite{chung2025theoretical}, and computer science~\cite{jimenez2023swe}, no standardized frameworks comprehensively assess quantum computing knowledge in these models. This absence is particularly concerning given the field's counterintuitive principles, and rapidly evolving terminology that challenge even domain experts. The complexity of quantum computing concepts, combined with their inherent mathematical abstraction, creates a particularly demanding test case for evaluating the depth of LLMs' specialized knowledge.
Without reliable evaluation metrics, LLMs risk spreading plausible but incorrect quantum information to educational and research communities, as hallucinations, reasoning errors, and factual inaccuracies have been widely documented in similarly complex and technically demanding specialized domains.~\cite{orgad2024llms, afane2025can,perkovic2024hallucinations}.

This creates an urgent need for robust quantum computing benchmarks as researchers, students, and industry professionals increasingly rely on these models for information and assistance with quantum tasks. The growing adoption of LLMs across academic institutions and quantum technology companies further amplifies the importance of ensuring these systems provide accurate information on this emerging field.  
To address these challenges, we present the following key contributions:
\begin{itemize}[noitemsep,topsep=0pt]
\item We assemble \textbf{2,000 multiple-choice questions}: 1,000 expert-written questions developed by quantum computing researchers and 1,000 questions extracted from research papers using LLMs and validated by domain experts, covering seven core topics including quantum algorithms, error correction, security protocols, distributed computing, quantum machine learning, gates and circuits, and foundational concepts.
\item We conduct extensive evaluation across \textbf{26 models}, comparing their performance against 43 quantum computing experts and practitioners to establish human baselines and assess how models perform relative to human capabilities across different experience levels.
\item We provide \textbf{multi-dimensional analysis} with an additional 350 open-ended questions requiring detailed explanations and 350 questions containing intentionally false premises to evaluate whether models can identify and correct erroneous assumptions embedded within the question's formulation (2,700 questions total), with a multilingual subset enabling cross-lingual evaluation in Spanish and French. Our analysis reveals systematic failures on advanced topics like quantum security despite strong performance on foundational concepts, alongside significant multilingual performance degradation and poor performance on detecting false premises.
\end{itemize}

\section{Related Work}

Despite significant advancements in both quantum computing and LLMs, their intersection remains surprisingly underexplored. Recent research has begun addressing this gap from different angles. Kashani \cite{kashani2024quantumllminstruct} introduced QuantumLLMInstruct (QLMMI), a dataset of over 500,000 instruction-problem pairs covering quantum cryptography, spin chain models, and Trotter-Suzuki decompositions. However, QLMMI's primary purpose is to enable instruction fine-tuning rather than comprehensive evaluation of quantum knowledge. While extensive in size, QLMMI relies entirely on synthetically generated content through a four-stage LLM pipeline. In contrast, Quantum-Audit offers 1,200 human-authored evaluation questions extracted directly from research literature published over four decades, prioritizing authentic scientific content over synthetic generation. Wang et al. \cite{wang2024grovergpt} introduced GroverGPT, an approach to simulating quantum algorithms using LLMs. Their 8-billion-parameter model is fine-tuned to approximate Grover's quantum search algorithm without explicitly representing quantum states. While GroverGPT demonstrates impressive capabilities in predicting specific quantum circuit outputs, it focuses exclusively on a single quantum algorithm rather than evaluating comprehensive knowledge across the quantum computing domain.

Complementary efforts have emerged focusing on quantum code generation and circuit implementation capabilities. Vishwakarma et al. \cite{vishwakarma2024qiskit} developed Qiskit HumanEval, a hand-curated benchmark of over 100 tasks designed to evaluate LLM performance in generating executable quantum code using the Qiskit SDK, complete with canonical solutions and comprehensive test cases. Guo et al. \cite{guo2025quanbench} introduced QuanBench, which evaluates quantum code generation across 44 programming tasks using both functional correctness (Pass@K) and quantum semantic equivalence (Process Fidelity) metrics, finding that current LLMs achieve below 40\% overall accuracy with frequent semantic errors including outdated API usage and incorrect algorithm logic. Yang et al. \cite{yang2024qcircuitnet} presented QCircuitNet, a large-scale hierarchical dataset for quantum algorithm design containing 120,290 data points with automatic syntax and semantic verification functions. At a lower abstraction level, Li et al. \cite{li2023qasmbench} developed QASMBench, a benchmark suite of low-level OpenQASM programs for evaluating NISQ devices and simulators. While these works provide valuable resources for assessing programming proficiency and implementation capabilities at various levels of quantum software development, they primarily target coding skills rather than evaluating deep conceptual understanding of quantum computing principles, algorithmic theory, or the ability to reason about quantum phenomena. Quantum-Audit addresses this by evaluating theoretical knowledge and conceptual understanding across quantum computing topics, from foundational algorithmic principles to advanced security protocols and attack vectors.

\begin{figure}[t]
  \centering
  \includegraphics[width=\columnwidth]{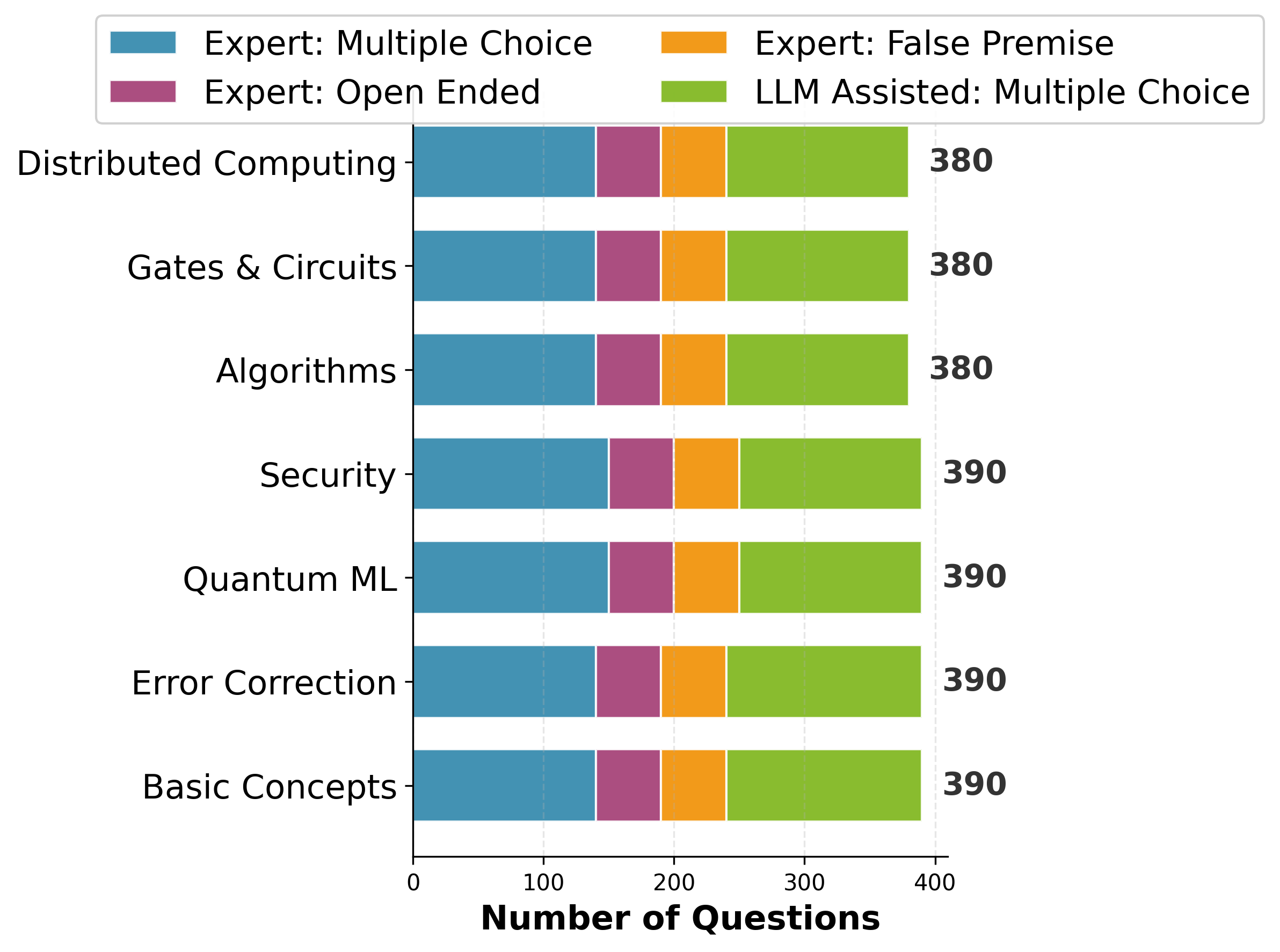}
  \caption{Distribution of the 2,700 benchmark questions by topic. Expert-written questions include multiple choice, open-ended, and false premise questions, while LLM-assisted questions consist solely of multiple choice. Topics are ordered by total question count to highlight relative coverage across different quantum computing domains.}
  \label{fig:question-distribution}
\end{figure}

\section{Quantum-Audit Dataset}

We constructed the Quantum-Audit dataset to evaluate quantum computing knowledge in LLMs across a wide range of topics and difficulty levels. The dataset includes 1,000 expert-written questions developed by quantum computing researchers. Beyond multiple-choice questions, we developed 350 open-ended questions requiring models to explain concepts without answer options, and 350 questions with false premises to assess models' ability to identify and correct faulty assumptions. A subset of 500 questions was translated into Spanish and French to assess cross-lingual generalization capabilities. To address concerns about model memorization, these questions reflect core concepts reformulated into original assessments rather than direct reproductions from source materials. This approach ensures that performance reflects genuine understanding rather than memorization of published text.

To expand our benchmark, we employed Gemini 3 Flash, GPT-4.1, and Claude Sonnet 4 to generate additional candidate questions using quantum computing literature as reference material. Different prompt engineering techniques were tested to optimize question generation quality. While zero-shot prompting produced acceptable results, few-shot prompting with five carefully selected examples from the existing subsets significantly improved the relevance and technical accuracy of generated questions. This approach generated over 8,000 candidate questions, subsequently filtered to remove low-quality or redundant items. The final selection included 1,000 high-quality questions validated by domain experts, bringing the total benchmark size to 2,700 questions. Figure \ref{fig:question-distribution} illustrates the distribution of these question types across different topics. Given the interconnected nature of quantum computing, some concepts naturally appear across multiple categories; for example, noise characterization relates to both Error Correction and Gates and Circuits, while variational optimization techniques bridge Quantum Algorithms and Quantum ML, as the same parameterized circuit methods underlie both algorithmic design and learning tasks. The multiple-choice format enables structured evaluation of quantum computing knowledge, open-ended questions assess performance when no answer options are given, and false premise questions evaluate whether models can recognize and correct incorrect assumptions from the user.\footnote{Dataset, code, and leaderboard available at \url{https://quantum-audit.github.io/}} We plan to update the leaderboard as new models are released to track advances in language model capabilities on quantum computing knowledge.

\section{Experiments}

We evaluated 26 LLMs using a consistent benchmarking pipeline. Closed-source models were accessed through their official APIs. This includes GPT models (GPT-5.2, GPT-5.2 Pro, GPT-5 mini, GPT-4.1, GPT-4.1 mini), Claude models (Opus 4.5, Sonnet 4.5, Sonnet 4, Haiku 4.5), and Gemini models (Gemini 3 Pro, Gemini 3 Flash, Gemini 2.5 Pro, Gemini 2.0 Flash-Lite). Open-access models were evaluated using either Hugging Face's Transformers library on a GPU cluster equipped with two Tesla V100 GPUs (32GB each) using FP16 inference, or Groq's API for faster and more efficient inference. This includes LLaMA models (ranging from 1B to 70B parameters), Microsoft Phi models (2.7B to 14.7B parameters), and Google Gemma models (2B to 9B parameters).

For experiment preparation, all benchmark questions were structured in JSON format for efficient processing and consistent evaluation across different model architectures. We developed standardized prompting templates for each question type to ensure fair comparison between models. This data preparation approach facilitated easier evaluation pipelines and ensured comparable results despite the diversity of model implementations and access methods. The benchmark includes multiple-choice questions, open-ended questions, and questions with false premises. To assess cross-lingual capabilities, a subset of questions was translated into Spanish and French. Key findings from these experiments are presented in the following tables and subsections.

\begin{table*}[htbp]
  \caption{Model accuracy on multiple-choice QA benchmark subsets. QA1000 Expert Written contains questions authored by quantum computing researchers, QA1000 LLM Extracted contains questions generated from literature using LLMs and validated by experts. QA2000 represents combined performance across both subsets. Top three models are highlighted in green, select top-performing open-source models are highlighted in blue, and bottom three models are highlighted in red.}
  \label{tab:qa_benchmark_results}
  \centering
  \renewcommand{\arraystretch}{1.25}
  \begin{tabular}{@{}llcccc@{}}
    \toprule
    & \textbf{Model} & \textbf{Provider} & \textbf{QA1000 Expert Written} & \textbf{QA1000 LLM Extracted} & \textbf{QA2000} \\
    \midrule
    \includegraphics[height=2ex]{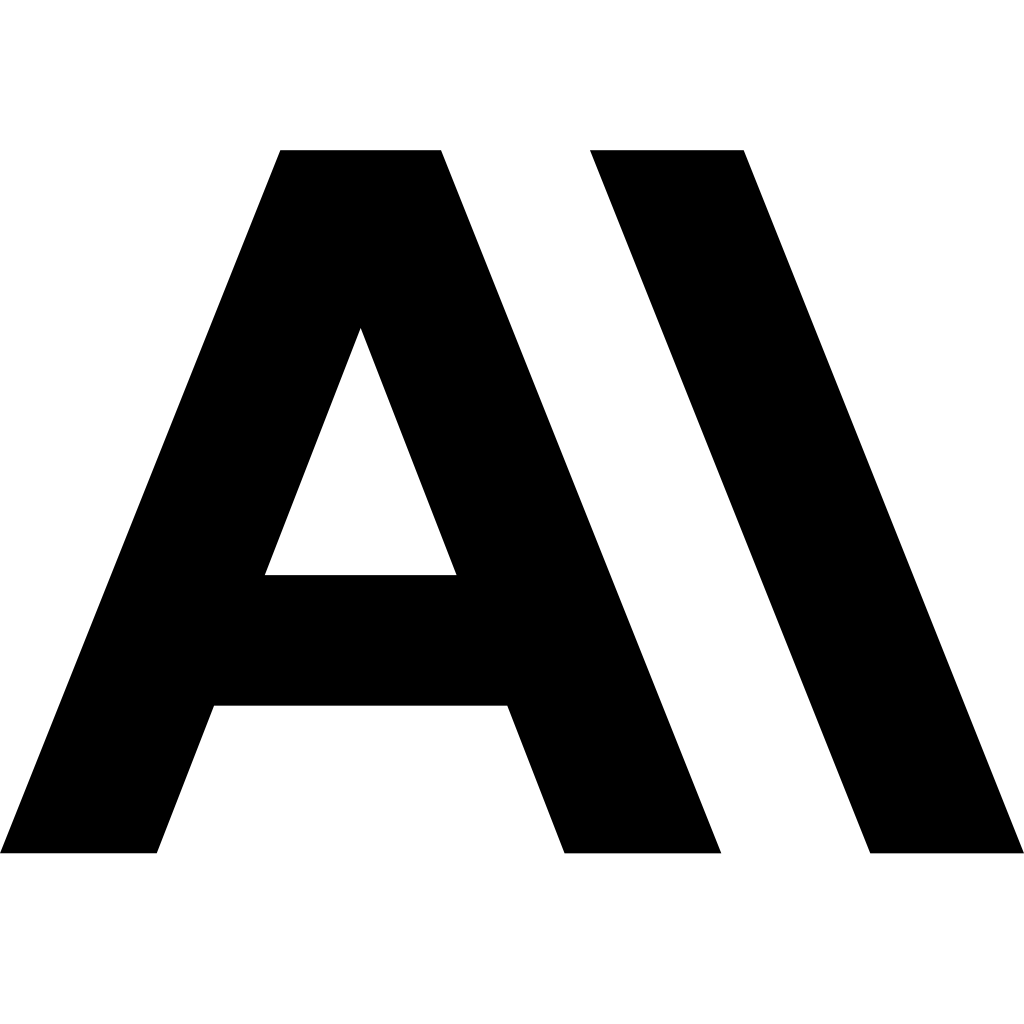} & \cellcolor{green!15}Claude Opus 4.5 & \cellcolor{green!15}Anthropic & \cellcolor{green!15}\textbf{78.40} & \cellcolor{green!15}89.60 & \cellcolor{green!15}\textbf{84.00} \\
    \includegraphics[height=2ex]{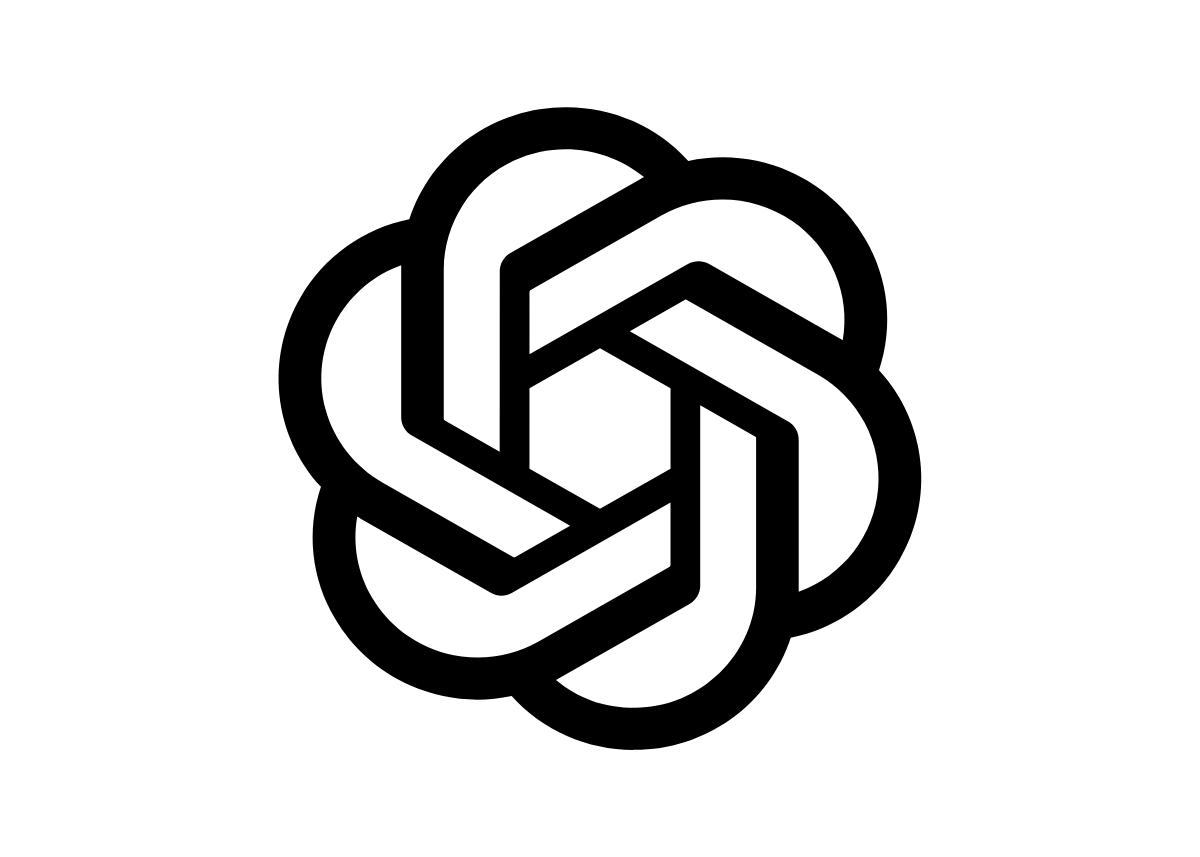} & \cellcolor{green!15}GPT-5.2 Pro & \cellcolor{green!15}OpenAI & \cellcolor{green!15}77.70 & \cellcolor{green!15}\textbf{89.80} & \cellcolor{green!15}83.75 \\
    \includegraphics[height=2ex]{logos/anthropic.png} & \cellcolor{green!15}Claude Sonnet 4.5 & \cellcolor{green!15}Anthropic & \cellcolor{green!15}77.20 & \cellcolor{green!15}89.40 & \cellcolor{green!15}83.30 \\
    \includegraphics[height=2ex]{logos/openai.png} & GPT-5.2 & OpenAI & 76.80 & 89.20 & 83.00 \\
    \includegraphics[height=2ex]{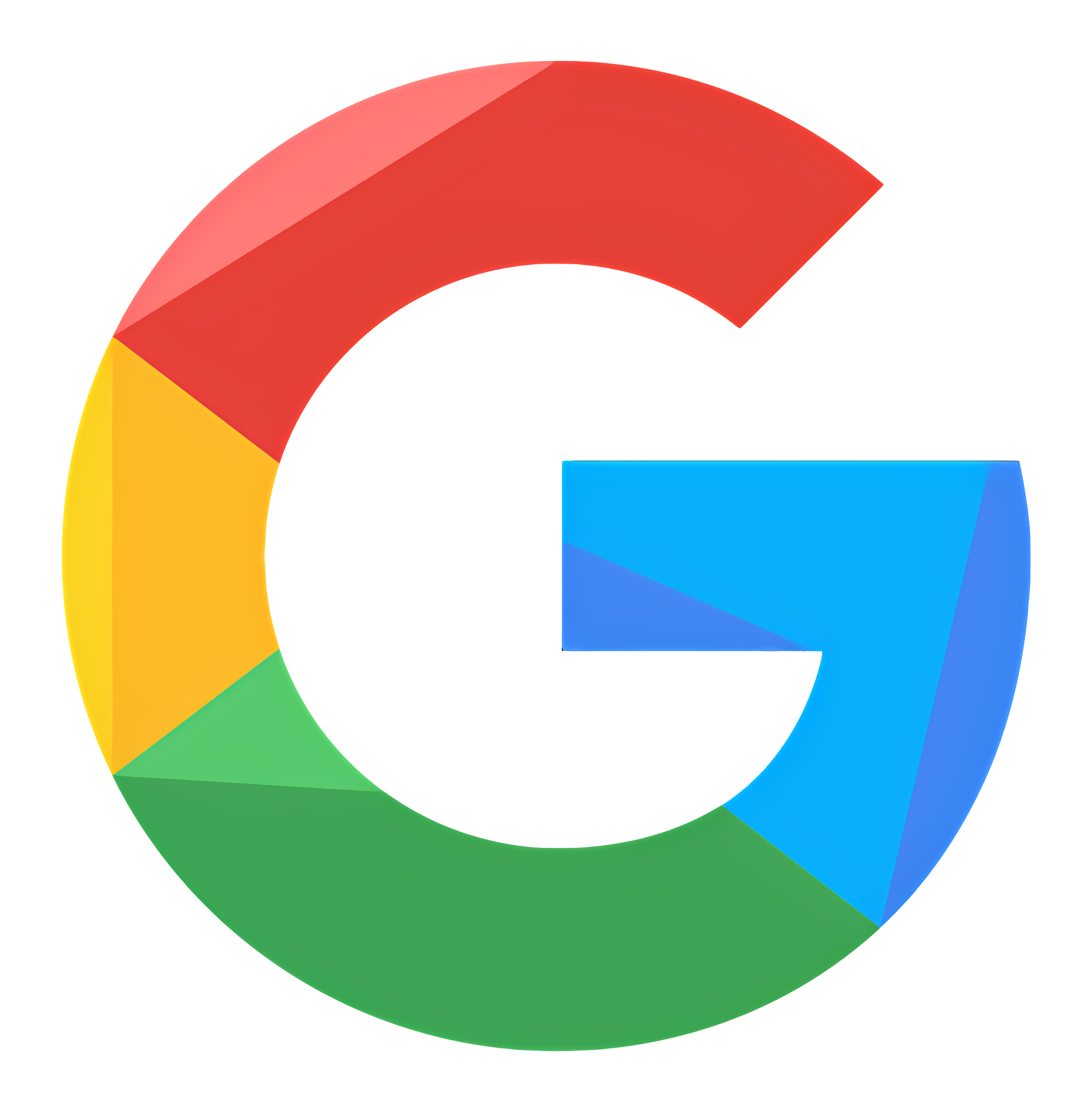} & Gemini 3 Pro & Google & 76.10 & 88.50 & 82.30 \\
    \includegraphics[height=2ex]{logos/anthropic.png} & Claude Sonnet 4 & Anthropic & 75.60 & 88.10 & 81.85 \\
    \includegraphics[height=2ex]{logos/google.png} & Gemini 3 Flash & Google & 74.90 & 87.60 & 81.25 \\
    \includegraphics[height=2ex]{logos/openai.png} & GPT-4.1 & OpenAI & 74.40 & 87.20 & 80.80 \\
    \includegraphics[height=2ex]{logos/google.png} & Gemini 2.5 Pro & Google & 73.70 & 86.70 & 80.20 \\
    \includegraphics[height=2ex]{logos/openai.png} & GPT-4.1 mini & OpenAI & 73.30 & 84.90 & 79.10 \\
    \includegraphics[height=2ex]{logos/openai.png} & GPT-5 mini & OpenAI & 71.50 & 85.80 & 78.65 \\
    \includegraphics[height=2ex]{logos/anthropic.png} & Claude Haiku 4.5 & Anthropic & 71.90 & 84.60 & 78.25 \\
    \includegraphics[height=2ex]{logos/google.png} & Gemini 2.0 Flash-Lite & Google & 71.10 & 84.30 & 77.70 \\
    \includegraphics[height=2ex]{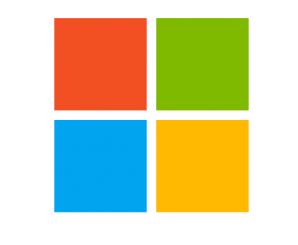} & \cellcolor{cyan!15}Phi-4-reasoning-plus & \cellcolor{cyan!15}Microsoft & \cellcolor{cyan!15}72.60 & \cellcolor{cyan!15}82.90 & \cellcolor{cyan!15}77.75 \\
    \includegraphics[height=2ex]{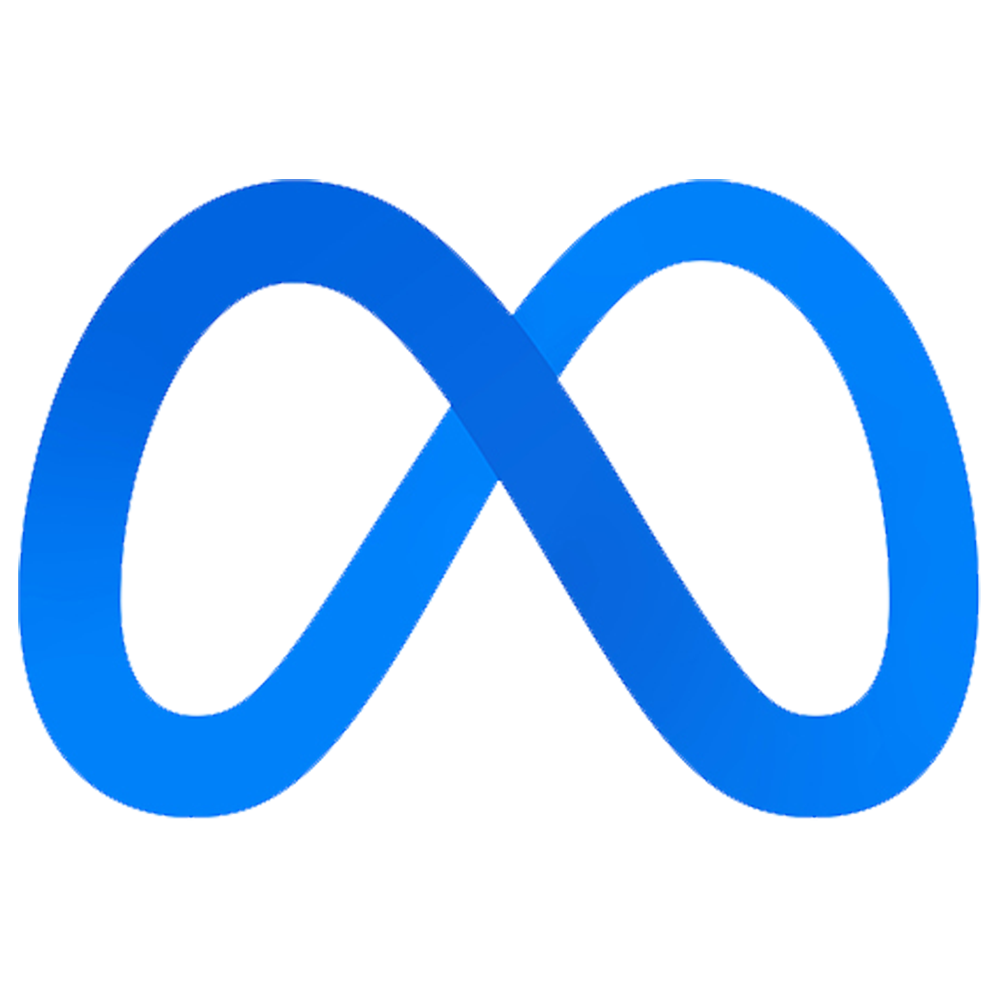} & \cellcolor{cyan!15}llama-3.3-70b-versatile & \cellcolor{cyan!15}Meta & \cellcolor{cyan!15}69.80 & \cellcolor{cyan!15}82.50 & \cellcolor{cyan!15}76.15 \\
    \includegraphics[height=2ex]{logos/meta.png} & \cellcolor{cyan!15}llama3-70b & \cellcolor{cyan!15}Meta & \cellcolor{cyan!15}68.40 & \cellcolor{cyan!15}82.80 & \cellcolor{cyan!15}75.60 \\
    \includegraphics[height=2ex]{logos/google.png} & gemma2-9b-it & Google & 67.10 & 79.90 & 73.50 \\
    \includegraphics[height=2ex]{logos/microsoft.png} & Phi-4-reasoning & Microsoft & 65.00 & 81.00 & 73.00 \\
    \includegraphics[height=2ex]{logos/meta.png} & Llama-3.1-8B-Instruct & Meta & 62.00 & 78.70 & 70.35 \\
    \includegraphics[height=2ex]{logos/meta.png} & Llama-3.1-8B & Meta & 57.30 & 75.90 & 66.60 \\
    \includegraphics[height=2ex]{logos/microsoft.png} & Phi-4-mini-reasoning & Microsoft & 58.60 & 72.80 & 65.70 \\
    \includegraphics[height=2ex]{logos/meta.png} & Llama-2-13b-chat-hf & Meta & 55.00 & 71.30 & 63.15 \\
    \includegraphics[height=2ex]{logos/google.png} & gemma-7b & Google & 52.80 & 70.00 & 61.40 \\
    \includegraphics[height=2ex]{logos/meta.png} & \cellcolor{red!10}Llama-3.2-1B-Instruct & \cellcolor{red!10}Meta & \cellcolor{red!10}50.40 & \cellcolor{red!10}68.40 & \cellcolor{red!10}59.40 \\
    \includegraphics[height=2ex]{logos/microsoft.png} & \cellcolor{red!10}phi-2 & \cellcolor{red!10}Microsoft & \cellcolor{red!10}42.40 & \cellcolor{red!10}60.90 & \cellcolor{red!10}51.65 \\
    \includegraphics[height=2ex]{logos/google.png} & \cellcolor{red!10}gemma-2-2b-it & \cellcolor{red!10}Google & \cellcolor{red!10}40.10 & \cellcolor{red!10}57.60 & \cellcolor{red!10}48.85 \\
    \bottomrule
  \end{tabular}
\end{table*}

\begin{figure*}[t]
\centering
\begin{tikzpicture}
\node[rounded corners=8pt, fill=gray!8, inner sep=10pt, text width=0.92\textwidth] {
\small
\textbf{Example Security Questions}\\[0.2cm]
\begin{itemize}[leftmargin=0.7cm, itemsep=0.15cm, topsep=0.1cm]
\item What specific attack technique can \colorbox{yellow!30}{manipulate the error rates of specific quantum gates}?
\item What specific vulnerability does a \colorbox{yellow!30}{quantum reorder attack} exploit?
\item What makes dynamical decoupling ineffective against \colorbox{yellow!30}{QubitHammer attacks}?
\end{itemize}
};
\end{tikzpicture}
\caption{Sample quantum security questions that demonstrate the technical specificity required for this topic area. Highlighted portions indicate concepts requiring knowledge of recent attack research where even leading models show reduced accuracy.}
\label{fig:security-questions}
\end{figure*}

\subsection{Comprehensive Model Evaluation on Core Benchmark and Across Topics}

Table~\ref{tab:qa_benchmark_results} reveals a clear difficulty difference between expert-written and LLM-extracted questions. Models consistently score lower on expert-written questions, with even top performers like Claude Opus 4.5 and GPT-5.2 Pro achieving only 77-78\% accuracy compared to nearly 90\% on LLM-extracted questions. This pattern holds across most models regardless of architecture or size. The gap suggests that expert-written questions demand deeper reasoning or test concepts that are less prominently featured in training data, while LLM-extracted questions may inadvertently align with the statistical patterns and phrasings models encountered during pre-training.

The topic-specific analysis in Table~\ref{tab:advanced_topics} shows where models struggle most. Leading models achieve over 90\% accuracy on Basic Concepts, demonstrating strong command of foundational quantum computing principles. Performance degrades on Quantum Algorithms, where top models drop to roughly 80-82\% accuracy. The sharpest decline appears in Quantum Security, where even the best models barely exceed 74\% and several fall below 70\%. Security questions prove particularly challenging because they address recent attack research, as illustrated in Figure~\ref{fig:security-questions}. These questions require knowledge of phase mismatch attacks that exploit hardware calibration vulnerabilities, crosstalk-based attacks that leverage multi-qubit interference, QubitHammer techniques that manipulate error rates through strategic gate scheduling, and quantum backdoor insertion methods. These topics represent rapidly evolving research areas where literature is limited and terminology less standardized. The performance pattern across model families suggests the challenge stems from the nature of cutting-edge quantum security research rather than model architecture.

\begin{table*}[htbp]
  \caption{Model accuracy on selected quantum topics. Accuracy above 80\% are shaded green and those below 50\% are shaded red.}
  \label{tab:advanced_topics}
  \centering
  \renewcommand{\arraystretch}{1.25}
  \begin{tabular}{@{}llcccc@{}}
    \toprule
    & \textbf{Model} & \textbf{Provider} & \textbf{Basic Concepts} & \textbf{Quantum Algorithms} & \textbf{Quantum Security} \\
    \midrule
    \includegraphics[height=2ex]{logos/openai.png} & GPT-5.2 Pro & OpenAI & \cellcolor{green!15}\textbf{93.10} & \cellcolor{green!15}\textbf{82.14} & 73.45 \\
    \includegraphics[height=2ex]{logos/anthropic.png} & Claude Opus 4.5 & Anthropic & \cellcolor{green!15}92.76 & \cellcolor{green!15}80.71 & 72.41 \\
    \includegraphics[height=2ex]{logos/anthropic.png} & Claude Sonnet 4.5 & Anthropic & \cellcolor{green!15}92.41 & 79.64 & \textbf{74.14} \\
    \includegraphics[height=2ex]{logos/google.png} & Gemini 3 Pro & Google & \cellcolor{green!15}91.72 & 78.21 & 71.38 \\
    \includegraphics[height=2ex]{logos/openai.png} & GPT-5.2 & OpenAI & \cellcolor{green!15}91.38 & \cellcolor{green!15}81.43 & 73.10 \\
    \includegraphics[height=2ex]{logos/anthropic.png} & Claude Sonnet 4 & Anthropic & \cellcolor{green!15}90.69 & 77.50 & 70.69 \\
    \includegraphics[height=2ex]{logos/google.png} & Gemini 3 Flash & Google & \cellcolor{green!15}83.45 & 76.43 & 72.76 \\
    \includegraphics[height=2ex]{logos/openai.png} & GPT-4.1 & OpenAI & \cellcolor{green!15}82.76 & 79.29 & 69.31 \\
    \includegraphics[height=2ex]{logos/google.png} & Gemini 2.5 Pro & Google & 79.66 & 75.71 & 68.97 \\
    \includegraphics[height=2ex]{logos/microsoft.png} & Phi-4-reasoning-plus & Microsoft & 78.62 & 76.79 & 67.24 \\
    \includegraphics[height=2ex]{logos/openai.png} & GPT-5 mini & OpenAI & 77.93 & 73.93 & 66.21 \\
    \includegraphics[height=2ex]{logos/anthropic.png} & Claude Haiku 4.5 & Anthropic & 77.24 & 74.64 & 70.34 \\
    \includegraphics[height=2ex]{logos/openai.png} & GPT-4.1 mini & OpenAI & 76.55 & 72.86 & 64.83 \\
    \includegraphics[height=2ex]{logos/google.png} & Gemini 2.0 Flash-Lite & Google & 75.86 & 73.21 & 65.17 \\
    \includegraphics[height=2ex]{logos/meta.png} & llama-3.3-70b-versatile & Meta & 74.48 & 75.36 & 69.66 \\
    \includegraphics[height=2ex]{logos/meta.png} & llama3-70b & Meta & 73.10 & 71.79 & 61.38 \\
    \includegraphics[height=2ex]{logos/google.png} & gemma2-9b-it & Google & 71.03 & 73.57 & 58.62 \\
    \includegraphics[height=2ex]{logos/microsoft.png} & Phi-4-reasoning & Microsoft & 69.31 & 71.43 & 56.90 \\
    \includegraphics[height=2ex]{logos/meta.png} & Llama-3.1-8B-Instruct & Meta & 67.59 & 69.29 & 62.41 \\
    \includegraphics[height=2ex]{logos/meta.png} & Llama-3.1-8B & Meta & 64.48 & 65.71 & 54.48 \\
    \includegraphics[height=2ex]{logos/microsoft.png} & Phi-4-mini-reasoning & Microsoft & 62.76 & 63.93 & \cellcolor{red!10}49.66 \\
    \includegraphics[height=2ex]{logos/meta.png} & Llama-2-13b-chat-hf & Meta & 60.69 & 59.64 & 51.72 \\
    \includegraphics[height=2ex]{logos/google.png} & gemma-7b & Google & 58.62 & 61.07 & \cellcolor{red!10}46.90 \\
    \includegraphics[height=2ex]{logos/meta.png} & Llama-3.2-1B-Instruct & Meta & 56.21 & 57.14 & \cellcolor{red!10}48.28 \\
    \includegraphics[height=2ex]{logos/microsoft.png} & phi-2 & Microsoft & \cellcolor{red!10}49.31 & 52.86 & \cellcolor{red!10}44.83 \\
    \includegraphics[height=2ex]{logos/google.png} & gemma-2-2b-it & Google & \cellcolor{red!10}48.97 & \cellcolor{red!10}48.93 & \cellcolor{red!10}41.38 \\
    \bottomrule
  \end{tabular}
\end{table*}

This performance gap between foundational and advanced topics is particularly revealing. The disparity suggests that models have absorbed well-documented principles but struggle with recent developments where literature is sparser. The consistency of this pattern across model families, indicates that the challenge lies in the nature of the material rather than individual model limitations.


\subsection{Performance Across Different Question Formats}

Our evaluation extended beyond multiple-choice questions to assess model capabilities across diverse cognitive demands. We developed 350 open-ended questions requiring models to explain quantum computing concepts without answer options, testing whether they can construct coherent explanations from understanding rather than recognizing correct answers. Additionally, we created 350 questions with intentionally false premises embedded in the question formulation itself. These questions test whether models can identify and correct erroneous assumptions rather than accepting faulty premises and attempting to answer anyway.

\begin{table}[htbp]
  \centering
  \caption{Model accuracy on open-ended and false premise questions out of 350 questions each}
  \label{tab:question-format-performance}
  \renewcommand{\arraystretch}{1.3}
  \begin{tabular}{@{}lcc@{}}
    \toprule
    \textbf{Model} & \textbf{Open-Ended (\%)} & \textbf{False Premise (\%)} \\
    \midrule
    GPT-5.2 Pro & \textbf{81.4} & 64.9 \\
    Claude Opus 4.5 & 79.7 & \textbf{65.7} \\
    Claude Sonnet 4.5 & 78.3 & 63.4 \\
    GPT-5.2 & 77.1 & 62.3 \\
    Gemini 3 Pro & 75.4 & 61.1 \\
    Claude Sonnet 4 & 74.9 & 60.6 \\
    Gemini 3 Flash & 72.6 & 58.9 \\
    GPT-4.1 & 71.1 & 57.7 \\
    Gemini 2.5 Pro & 64.4 & 56.3 \\
    \bottomrule
  \end{tabular}
\end{table}

\begin{figure*}[t]
\centering
\begin{tikzpicture}
\node[rounded corners=8pt, fill=gray!8, inner sep=12pt, text width=0.95\textwidth] {
\normalsize
\textbf{Example 1: Mischaracterizing Algorithm Capabilities}\\[0.3cm]
\textit{Question:} Given that \colorbox{yellow!30}{Shor's algorithm provides exponential speedup for all NP-complete problems}, which optimization technique yields the best performance in practice?\\[0.3cm]
\textit{Expected Response:} Reject the premise. Shor's algorithm specifically addresses integer factorization and discrete logarithm problems, not NP-complete problems generally.\\[0.5cm]

\textbf{Example 2: Incorrect Defense Mechanisms}\\[0.3cm]
\textit{Question:} Since \colorbox{yellow!30}{dynamical decoupling is the primary defense against QubitHammer attacks}, what additional protection layer should be implemented to ensure complete security?\\[0.3cm]
\textit{Expected Response:} Reject the premise. Dynamical decoupling is actually ineffective against QubitHammer attacks, which manipulate error rates through strategic gate scheduling.
};
\end{tikzpicture}
\caption{Examples of false premise questions where models must identify and correct erroneous assumptions embedded in the question formulation. Highlighted portions indicate the false premises that should be rejected rather than accepted.}
\label{fig:false-premise-examples}
\end{figure*}

For false premise questions, the correct response requires rejecting the question's underlying assumption. Figure~\ref{fig:false-premise-examples} illustrates two examples. In the first, the question assumes Shor's algorithm solves all NP-complete problems, when it actually addresses specific problems like integer factorization. In the second, the question assumes dynamical decoupling defends against QubitHammer attacks, when it is actually ineffective against them. Models that simply answer these questions as posed demonstrate a concerning tendency to build upon incorrect user assumptions rather than challenging them. We evaluated nine leading models across these formats: Claude Opus 4.5, Sonnet 4.5, Sonnet 4, GPT-5.2 Pro, GPT-5.2, GPT-4.1, Gemini 3 Pro, 3 Flash, and 2.5 Pro. 

Table~\ref{tab:question-format-performance} presents the results. Models performed well on open-ended questions, with top performers reaching above 80\% accuracy, demonstrating they can generate coherent explanations when the task is clearly defined. However, performance dropped notably on false premise questions, where even the best models achieved only around 65\% accuracy. This gap reveals that models frequently accept faulty assumptions rather than questioning them, potentially misleading users who rely on these systems for technical guidance.

\subsection{Performance with Agentic and Deep Research Modes}

To explore whether models could improve their quantum computing knowledge through enhanced reasoning capabilities, we evaluated frontier models equipped with agentic and deep research modes. These capabilities enable models to perform multi-step reasoning, search external sources, and gather information across multiple queries without requiring additional training data.

We evaluated frontier models equipped with their respective advanced capabilities: Claude Opus 4.5 Research Mode, Claude Sonnet 4.5 Research Mode, GPT-5.2 Deep Research, GPT-5.2 Agent Mode, and Gemini 3 Deep Research. These modes allow models to break down complex questions, search for relevant information, and construct answers through extended reasoning processes. We tested these capabilities on a 500-question subset, which provides a balanced evaluation across all quantum computing topics in our benchmark.

\begin{table}[htbp]
\centering
\caption{Performance of frontier models with advanced reasoning capabilities on 500-question subset, ordered by improvement. Average improvement is 6.7 percentage points.}
\label{tab:agentic_modes}
\renewcommand{\arraystretch}{1.3}
\begin{tabular}{@{}lccc@{}}
\toprule
\textbf{Model} & \textbf{Before} & \textbf{After} & \textbf{$\Delta$} \\
\midrule
Gemini 3 (Deep Research) & 76.2 & 84.8 & +8.6 \\
Claude Opus 4.5 (Research) & 78.4 & 85.6 & +7.2 \\
GPT-5.2 (Agent Mode) & 76.8 & 83.4 & +6.6 \\
Claude Sonnet 4.5 (Research) & 77.2 & 83.0 & +5.8 \\
GPT-5.2 (Deep Research) & 76.8 & 82.2 & +5.4 \\
\bottomrule
\end{tabular}
\end{table}

Table~\ref{tab:agentic_modes} shows the performance of these advanced reasoning modes, ordered by improvement magnitude. Gemini 3 Deep Research showed the largest gain at 8.6 percentage points, reaching 84.8\% accuracy. Claude Opus 4.5 with Research Mode achieved the highest final score at 85.6\% with a 7.2 point improvement. GPT-5.2 Agent Mode gained 6.6 points to reach 83.4\%, while Claude Sonnet 4.5 Research Mode and GPT-5.2 Deep Research showed improvements of 5.8 and 5.4 points respectively. The average improvement across all models was 6.7 percentage points, demonstrating meaningful gains from these advanced capabilities. These results indicate that agentic and deep research modes provide notable benefits for quantum computing knowledge assessment. The improvements suggest these modes help models access and integrate information more effectively, though none exceeded 90\% accuracy. Advanced reasoning modes offer key advantages: (1) they require no training data or computational resources, (2) they can access current information beyond the model's knowledge cutoff, and (3) they can be applied to the largest and most capable models.

\begin{figure*}[h!] 
\centering 
\includegraphics[width=1\textwidth]{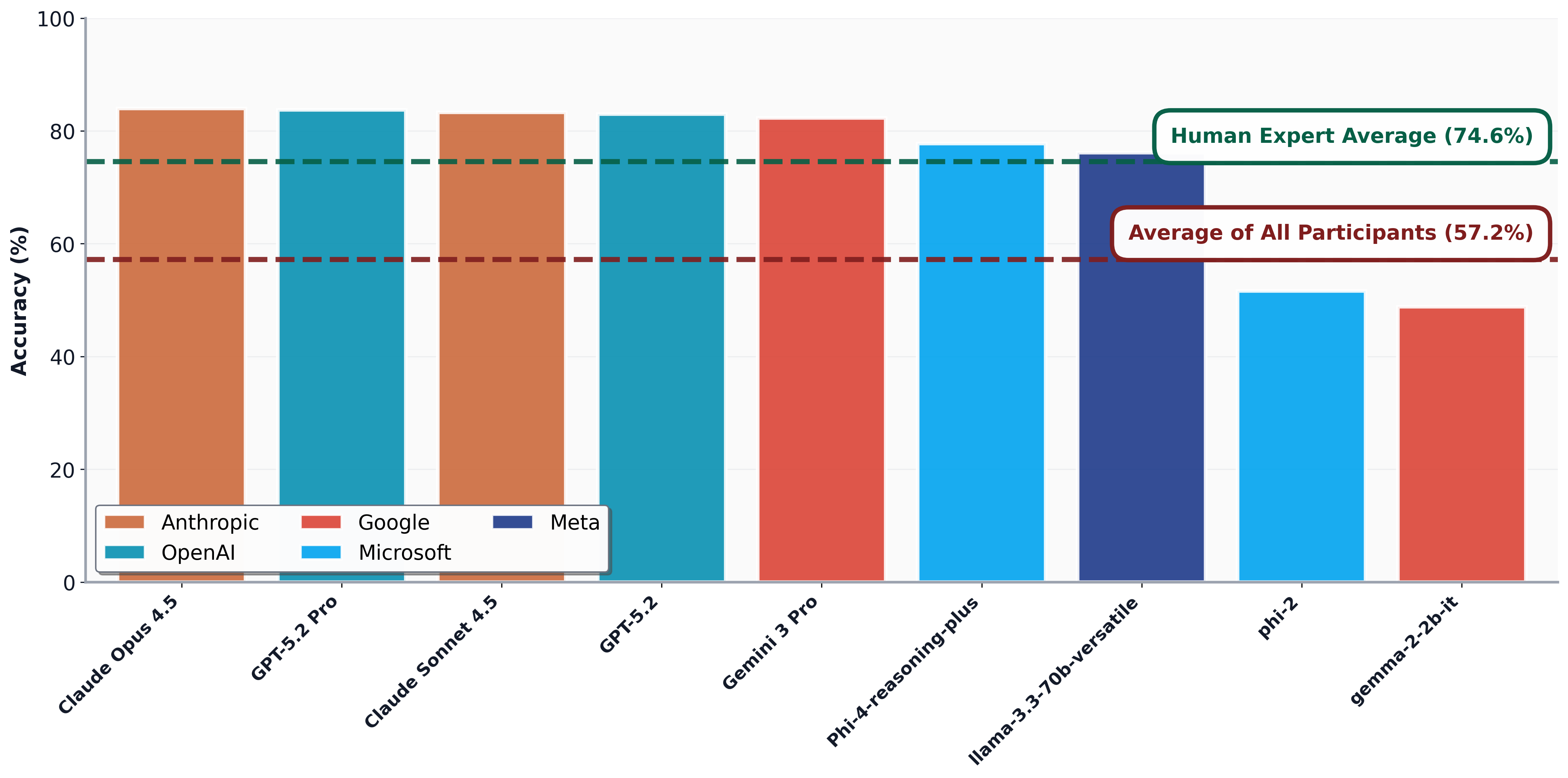} 
\caption{Performance comparison of selected LLMs across different capability tiers on the QA2000 benchmark against human baselines. The visualization includes 9 representative models ranging from top performers to those scoring below novice human levels. Bars are colored by model provider.} 
\label{fig:sm-fig1} 
\end{figure*}

\subsection{Human Performance Baseline Study}

To establish a human baseline for comparison with language model performance, we conducted a survey study with quantum computing researchers and practitioners. We carefully selected 30 questions from Quantum-Audit covering different topic areas and complexity levels to assess human expertise across the quantum computing domain. The survey included questions from all seven categories. Participants were recruited from academic institutions and quantum computing research groups. Each respondent provided background information including their highest education level, years of experience in quantum computing, and age group. Further details on each participant's background and individual score are provided in the appendix, offering context for the distribution shown here. The sample questions below illustrate the style and difficulty of the survey items used in this comparison.

The majority of models shown exceed the all-participants average of 57.2\%, while several surpass the expert average of 74.6\%. Figure~\ref{fig:sm-fig1} highlights the performance range in quantum computing capabilities, from leading models like Claude Opus 4.5 (84.00\%) and GPT-5.2 Pro (83.75\%) to models performing well below the participant average, such as phi-2 (51.65\%) and gemma-2-2b-it (48.85\%). This selection demonstrates that quantum computing proficiency varies across model families, sizes, and providers.

\begin{figure}[!t]
\centering
  \includegraphics[width=0.5\textwidth]{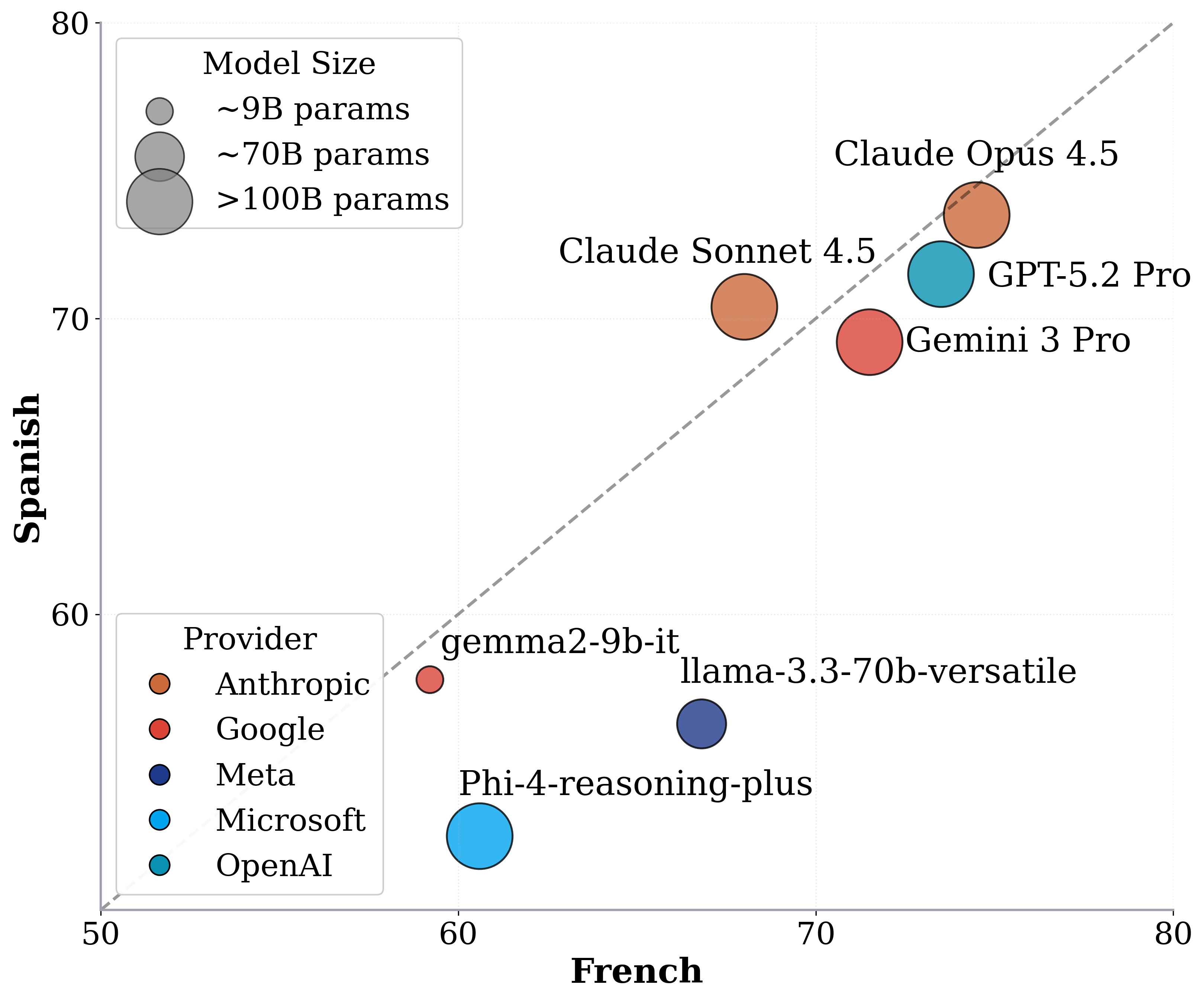}
\caption{Bubble chart of Spanish (horizontal) versus French (vertical) accuracy on the QA500 benchmark. Each bubble's area is proportional to the model's parameter count; colors indicate providers. The diagonal dashed line marks equal performance across the two languages. Bubbles below the line signal larger accuracy loss in Spanish.}
\label{fig:multilingual-performance}
\vspace{0.5cm}
\end{figure}

\subsection{Multilingual Benchmark Performance}
To investigate how quantum computing knowledge transfers across languages, we evaluated all models on Spanish and French translations of QA500. This experiment provides quantitative insights into linguistic generalization of specialized technical knowledge. Figure \ref{fig:multilingual-performance} shows Spanish versus French accuracy for selected models. The distribution reveals systematic language-dependent performance gaps, with models consistently performing better in French than Spanish. Among the top-performing models (Claude Opus 4.5, GPT-5.2 Pro, Claude Sonnet 4.5, Gemini 3 Pro), French accuracy averages 69.4\%, while Spanish accuracy averages 71.0\%, though individual results show variation with some models performing better in Spanish. The most linguistically robust models (Claude Opus 4.5, GPT-5.2 Pro) maintain accuracy above 72\% in both languages, while smaller models like Phi-4-reasoning-plus and gemma2-9b-it show more pronounced degradation. The diagonal reference line in Figure \ref{fig:multilingual-performance} highlights the performance relationship between the two languages across models of varying sizes and capabilities.

\section{Discussion}

Our evaluation reveals a clear performance pattern across all tested models: strong results on foundational topics with significant decline on advanced domains. Top models achieve over 92\% accuracy on basic quantum concepts but drop below 75\% on quantum security questions. This performance drop is particularly evident in questions addressing hardware-based vulnerabilities and side-channel attack vectors, where even the most advanced models struggle to provide accurate responses consistently. The results highlight a widening gap between state-of-the-art LLMs and smaller models, with high-capacity proprietary model showing clear advantages across all evaluation formats.

The expert-written versus LLM-extracted question comparison reveals an important pattern. Models score 10-15 percentage points lower on expert-written questions compared to LLM-extracted ones, even among top performers. This gap suggests that expert-written questions demand deeper reasoning or test concepts less prominently featured in training data, while LLM-extracted questions may inadvertently align with statistical patterns models encountered during pre-training. The consistency of this pattern across model families indicates that the challenge lies in the nature of the material rather than individual architectural limitations.

Question format analysis demonstrates critical differences in model capabilities. While leading models maintain strong performance on open-ended questions (GPT-5.2 Pro at 81.4\%, Claude Opus 4.5 at 79.7\%), all models show substantial degradation on false premise questions. Even the best performers achieve only around 65\% accuracy when required to identify and reject faulty assumptions embedded in questions. This reveals a concerning tendency: models frequently accept and work within erroneous premises rather than challenging them, behavior that could mislead users relying on these systems for technical guidance. Our evaluation of agentic and deep research modes shows promising results, with an average improvement of 6.7 percentage points when models can access external information and perform multi-step reasoning. Gemini 3 showed the largest gain (8.6 points), while Claude Opus 4.5 achieved the highest final accuracy (85.6\%). These capabilities offer practical advantages: they require no training data, can access current information beyond knowledge cutoffs, and apply to the largest models. Multilingual evaluation reveals systematic performance gaps across languages, with top-performing models (Claude Opus 4.5, GPT-5.2 Pro) maintaining above 72\% accuracy in both Spanish and French, while smaller models show more pronounced degradation.

\section{Limitations and Future Work}

Our evaluation relies primarily on accuracy as the central performance metric. We chose accuracy because it directly measures factual correctness, provides clear interpretability, and applies consistently across all question formats in our benchmark. For open-ended questions, responses are evaluated based on whether they contain the correct information and reasoning, making accuracy an appropriate measure of knowledge rather than writing style. For false premise questions, accuracy captures whether models correctly identify and reject faulty assumptions, which is precisely the capability we aim to assess. While alternative metrics like calibration scores or semantic similarity measures could complement our analysis, accuracy remains the most direct and interpretable measure for evaluating quantum computing knowledge across diverse question types. Future work includes expanding multilingual coverage beyond the current 500-question Spanish and French subset, adding additional languages, and increasing the diversity of non-English source materials to provide a more complete view of cross-lingual performance.

\section{Conclusion}

As Large Language Models are increasingly used for quantum computing education and research, rigorous domain evaluation becomes essential. Quantum-Audit provides comprehensive assessment with 2,000 multiple-choice questions, 350 open-ended questions, and 350 false premise questions across seven core domains. We evaluate 26 models and establish human baselines through surveys with 43 quantum computing experts and practitioners.

Our findings reveal consistent patterns: top models achieve over 92\% on basic concepts yet fall below 75\% on quantum security questions addressing recent attack research. Expert-written questions prove harder than LLM-extracted ones, with models scoring 10-15 percentage points lower even among top performers. Question format significantly impacts performance: leading models maintain strong results on open-ended questions (GPT-5.2 Pro at 81.4\%, Claude Opus 4.5 at 79.7\%) but all models degrade on false premise questions, with even the best achieving only 65\% accuracy. This reveals models' tendency to accept faulty assumptions rather than challenge them. 

Relative to human baselines (23\%--86\%), top models exceed the expert average of 74.6\%, with Claude Opus 4.5 reaching 84\% overall accuracy. Agentic and deep research modes show promise with an average improvement of 6.7 percentage points, though no model exceeded 86\% on the QA500 subset even with external information access. Multilingual testing reveals performance variations, with top models maintaining above 72\% in both Spanish and French while smaller models show pronounced degradation. These results demonstrate both the progress and remaining challenges in specialized quantum computing knowledge, highlighting the importance of continued evaluation as the field evolves.

\bibliographystyle{ACM-Reference-Format}
\bibliography{references}

\newpage

\onecolumn

\section{Appendices}

\subsection{Benchmark Development and Dataset Access}

We provide public access to the Quantum-Audit benchmark through an interactive platform at \url{https://quantum-audit.github.io/}. 

\begin{figure*}[h!] 
\centering 
\includegraphics[width=1\textwidth]{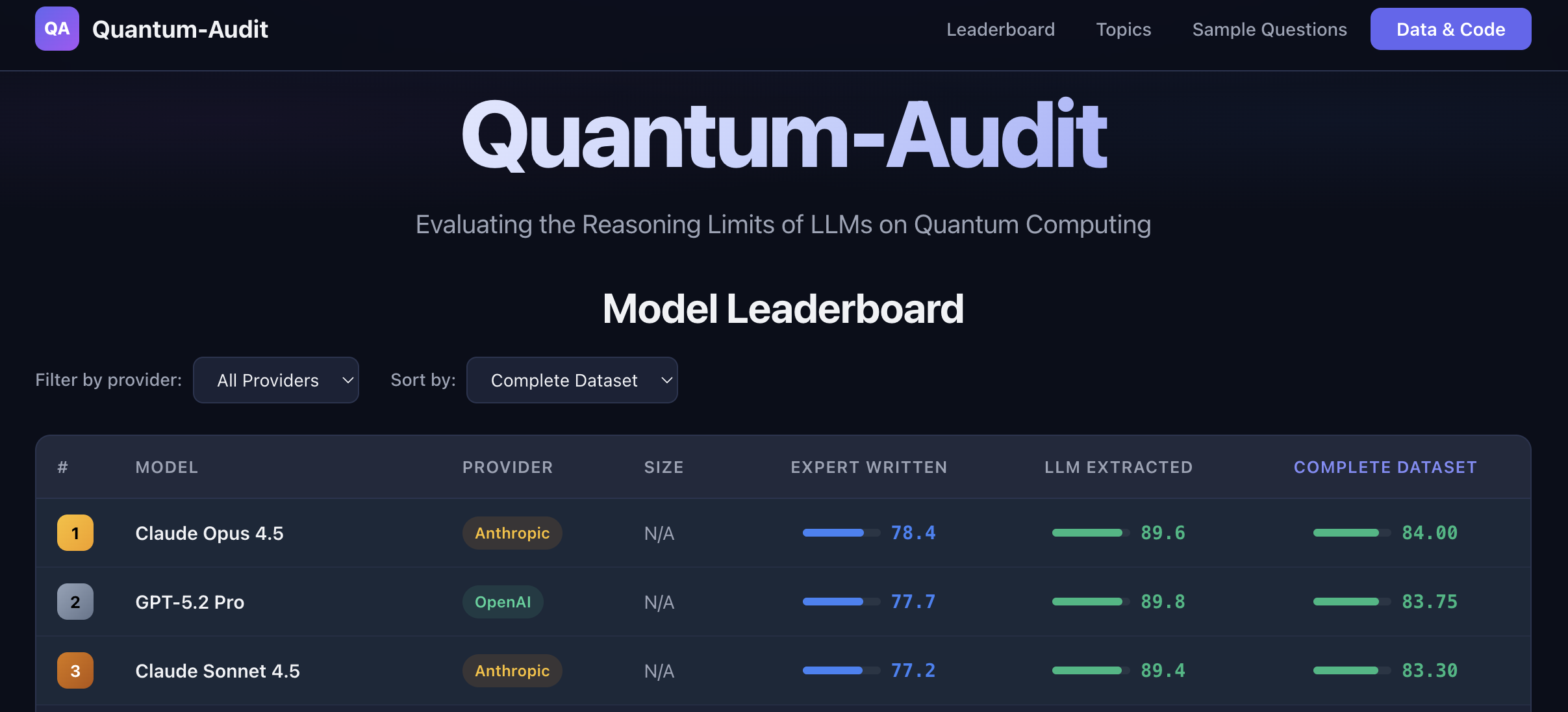} 
\caption{Quantum-Audit platform interface showing the model leaderboard and direct access to benchmark datasets and evaluation code.} 
\label{fig:platform} 
\end{figure*}

Figure~\ref{fig:platform} shows the platform interface, which includes a model leaderboard and direct links to download the complete dataset and evaluation code. Researchers can access all benchmark questions, view detailed model comparisons, and obtain the complete codebase for reproducing our experiments.

The question development process involved careful extraction and reformulation of concepts from quantum computing literature. For expert-written questions, domain researchers with specialized knowledge in quantum computing developed original assessments targeting core concepts, algorithmic principles, and theoretical foundations. Each question underwent multiple rounds of verification to ensure correctness and appropriate difficulty. For LLM-extracted questions, language models generated candidate questions from over 200 research papers published between 1980 and 2026, with domain experts filtering and validating each suggestion to maintain consistency with the benchmark's quality standards. This combined approach produced questions covering foundational topics to cutting-edge research areas, with answer options developed to test conceptual understanding rather than surface-level recall. Below we illustrate this process with an example:

\begin{tcolorbox}[colback=gray!5, colframe=gray!40, title=Example: Question Development Process, rounded corners]
\textbf{Initial concept from paper:} Quantum circuit synthesis involves decomposing unitary operations into implementable gate sequences.

\textbf{Generated question:} What is the primary purpose of quantum circuit synthesis?

\textbf{Initial answer options (6 generated):}
\begin{itemize}
  \item[A.] To convert a quantum circuit into a classical circuit by removing superposition properties
  \item[B.] To merge multiple unitary matrices into a single high-dimensional operator without gate decomposition
  \item[C.] To decompose a unitary matrix representing the circuit into a sequence of gates from the native gate set
  \item[D.] To encode classical information into qubit states without performing any gate-level modifications
  \item[E.] To simulate quantum circuits on classical computers using tensor networks
  \item[F.] To optimize quantum algorithms for specific hardware architectures
\end{itemize}

\textbf{Final selection (4 options):}
Options E and F were eliminated as they describe related but distinct processes. The final question includes the correct answer (C) and three plausible distractors that test understanding of quantum circuit concepts.
\end{tcolorbox}

\begin{tcolorbox}[colback=gray!5, colframe=gray!40, title=Quantum Circuit Synthesis, rounded corners]
\textbf{Question:} What is the primary purpose of quantum circuit synthesis?
\begin{itemize}
  \item[A.] To convert a quantum circuit into a classical circuit by removing superposition properties
  \item[B.] To merge multiple unitary matrices into a single high-dimensional operator without gate decomposition
  \item[C.] To decompose a unitary matrix representing the circuit into a sequence of gates from the native gate set
  \item[D.] To encode classical information into qubit states without performing any gate-level modifications
\end{itemize}
\textbf{Answer:} C
\end{tcolorbox}

\subsection{Question Filtering and Quality Control}

The automated question generation process required rigorous filtering to ensure benchmark quality. From the initial pool of over 8,000 candidate questions, we systematically removed those that tested adjacent domains rather than quantum computing concepts. The filtering criteria focused on eliminating questions that, while potentially appearing in quantum computing literature, primarily assessed knowledge outside the field's core competencies. Examples of filtered questions include:

\begin{tcolorbox}[colback=purple!5, colframe=purple!50!black, title=Example: Filtered Question - General Cybersecurity, rounded corners]
\textbf{Question:} Which encryption protocol is most commonly used for securing HTTP connections?
\begin{itemize}
  \item[A.] TLS/SSL
  \item[B.] SSH
  \item[C.] IPSec
  \item[D.] WPA2
\end{itemize}
\textit{Reason for filtering:} While encryption is relevant to quantum cryptography, this question addresses classical network security without quantum computing connection.
\end{tcolorbox}

\begin{tcolorbox}[colback=purple!5, colframe=purple!50!black, title=Example: Filtered Question – Mathematical Modeling, rounded corners]
\textbf{Question:} In the analysis of an ordinary differential equation system, what does a non-positive log-norm of the coefficient matrix imply?
\begin{itemize}
\item[A.] The system is unstable for all inputs
\item[B.] The matrix has only imaginary eigenvalues
\item[C.] The solution decays or remains bounded over time
\item[D.] The matrix is diagonalizable over the complex field
\end{itemize}
\textit{Reason for filtering:} While this concept appears in resource analyses for quantum-inspired algorithms, it tests classical stability theory in differential equations and does not assess quantum computing knowledge.
\end{tcolorbox}

After removing duplicate questions, filtering irrelevant content, and conducting manual quality review, we retained 1,000 high-quality questions. For each retained question, we selected the four most relevant answer options from the initial six, ensuring each question had one correct answer and three well-crafted distractors that effectively test quantum computing knowledge.

\subsection{Question Translation}

For the 500-question multilingual subset, we created Spanish and French translations using Gemini 3 Flash, GPT-4.1, and Claude Sonnet 4 to generate initial translations. A typical translation prompt was structured as follows:

\begin{verbatim}
"Translate the following quantum computing question from English to French, 
maintaining technical accuracy and appropriate scientific terminology:
[Question and answer options]"
\end{verbatim}

For each question, we collected translations from all models and selected the most accurate version based on technical terminology and conceptual accuracy. The translation process preserved the semantic content while adapting to language-specific conventions for scientific terminology.

\newpage
Examples of translated questions include:

\begin{tcolorbox}[colback=cyan!5, colframe=cyan!50!black, title=French Translation Example 1, rounded corners]
\textbf{Question:} Pourquoi les attaques par impulsion à grande échelle sont-elles difficiles à réaliser dans les systèmes partagés ?
\begin{itemize}
  \item[A.] Elles dépendent d'un accès chiffré aux qubits
  \item[B.] Elles nécessitent un accès à la machine au niveau administrateur
  \item[C.] Elles requièrent de nombreux qubits, auxquels les utilisateurs n'ont généralement pas accès
  \item[D.] Elles échouent si la machine est calibrée
\end{itemize}
\textbf{Answer:} C
\end{tcolorbox}

\begin{tcolorbox}[colback=cyan!5, colframe=cyan!50!black, title=French Translation Example 2, rounded corners]
\textbf{Question:} Pourquoi les algorithmes quantiques paramétriques sont-ils difficiles à vérifier sémantiquement ?
\begin{itemize}
  \item[A.] Ils utilisent des paramètres fixes définis dans le matériel
  \item[B.] Ils reposent uniquement sur un post-traitement classique
  \item[C.] Leurs paramètres entraînés manquent d'interprétabilité inhérente
  \item[D.] Leur structure est identique pour tous les ensembles de données
\end{itemize}
\textbf{Answer:} C
\end{tcolorbox}

\begin{tcolorbox}[colback=orange!10!white, colframe=orange!30!brown, title=Spanish Translation Example 1, rounded corners]
\textbf{Question:} ¿Cuál es la principal diferencia entre la privacidad diferencial clásica y la privacidad diferencial cuántica?
\begin{itemize}
  \item[A.] La PD cuántica extiende las garantías de privacidad a estados cuánticos indistinguibles utilizando distancias de traza
  \item[B.] La PD cuántica elimina la necesidad de análisis probabilístico
  \item[C.] La PD cuántica se aplica solo a registros de qubits entrelazados
  \item[D.] La PD cuántica se impone eliminando los resultados de medición de qubits
\end{itemize}
\textbf{Answer:} A
\end{tcolorbox}

\begin{tcolorbox}[colback=orange!10!white, colframe=orange!30!brown, title=Spanish Translation Example 2, rounded corners]
\textbf{Question:} ¿Qué algoritmo clásico se utiliza comúnmente después del paso cuántico del Algoritmo de Shor?
\begin{itemize}
  \item[A.] Algoritmo de Dijkstra
  \item[B.] Expansión de fracciones continuas
  \item[C.] Integración de Monte Carlo
  \item[D.] Búsqueda binaria
\end{itemize}
\textbf{Answer:} B
\end{tcolorbox}

\subsection{Question Format Diversification}

We developed 350 open-ended questions and 350 false premise questions to evaluate different cognitive capabilities. Open-ended questions assess whether models can construct explanations without answer options, while false premise questions test whether models can identify and reject faulty assumptions embedded in question formulations.

\begin{tcolorbox}[colback=gray!5, colframe=gray!40, title=False Premise Question Examples, rounded corners]
\textbf{Question:} Given that DRAG pulse shaping completely eliminates all unwanted transitions to higher energy levels in multi-qubit systems, what is the optimal pulse duration for achieving maximum gate fidelity?

\textit{\textbf{Expected Response:}} Reject the premise. While DRAG pulse shaping addresses unwanted coupling to higher energy levels, it does not completely eliminate all such transitions. It reduces leakage errors but cannot fully prevent them.

\vspace{0.3cm}

\textbf{Question:} Since geometric phase gates in trapped ion quantum computing rely on direct Coulomb interaction between physically adjacent ions, what minimum ion separation distance is required for gate operations?

\textit{\textbf{Expected Response:}} Reject the premise. Geometric phase gates in trapped ion systems use spin-dependent optical forces that drive ions along closed trajectories in phase space, not direct Coulomb interaction between ions.
\end{tcolorbox}

Open-ended questions were designed to assess deeper understanding and explanatory capabilities:

\begin{tcolorbox}[colback=gray!5, colframe=gray!40, title=Open-Ended Question Examples, rounded corners]
\textbf{Question:} What enables the implementation of controlled-SWAP (Fredkin) gates in linear optical quantum computing?

\vspace{0.3cm}

\textbf{Question:} What is the quantum mutual information used for in the analysis of quantum circuits?

\vspace{0.3cm}

\textbf{Question:} What is the key advantage of expressing quantum circuits in the phase polynomial formalism?

\vspace{0.3cm}

\textbf{Question:} What is the central challenge in implementing the exponential SWAP operation directly in quantum hardware?
\end{tcolorbox}

For each open-ended question, we developed sample answers to facilitate consistent evaluation across different models. These questions were assessed to determine whether model responses captured the essential concepts and technical accuracy required for each topic.

\subsection{Human Performance Baseline Study}

Table~\ref{tab:human-participants} reports accuracies for the first 20 respondents. Scores range from 23.3\% to 86.7\%, with an overall average of 57.2\%. Participants with 5+ years of experience achieved an average of 79.4\%, providing a reference point for expert-level performance. Education level shows a clear pattern: all PhD-trained participants scored at or above 73.3\%, while no BS-level participant reached 60\%. These results provide a concrete reference distribution for interpreting model-human comparisons in the main results.

\subsection*{References for Each Topic}
Table~\ref{tab:sources} lists the literature sources and citation coverage for all seven benchmark topics. Rapidly developing areas such as quantum cybersecurity and quantum machine learning rely heavily on the most recent papers to capture ongoing advances, whereas foundational categories such as quantum theory and quantum error correction draw on a broader historical record to reflect the principles that remain central to the discipline. This distribution ensures that the benchmark balances up-to-date research with enduring theoretical foundations, giving a clear view of how source material supports each topic area.

\begin{table}[htbp]
\caption{Human participant survey results on 30-question quantum computing assessment}
\label{tab:human-participants}
\centering
\small
\renewcommand{\arraystretch}{1.25}
\begin{tabular}{@{}lccccc@{}}
\toprule
\textbf{Participant} & \textbf{Education} & \textbf{Experience} & \textbf{Age Group} & \textbf{Score} & \textbf{Accuracy} \\
\midrule
P1 & MS  & 2--5 yrs     & 25--35 & 19/30 & 63.3\% \\
P2 & BS  & \textless 1 yr & 18--25 & 14/30 & 46.7\% \\
P3 & PhD & 5+ yrs       & 35--45 & 25/30 & 83.3\% \\
P4 & MS  & 1--2 yrs     & 25--35 & 21/30 & 70.0\% \\
P5 & PhD & 2--5 yrs     & 35--45 & 23/30 & 76.7\% \\
P6 & BS  & \textless 1 yr & 18--25 & 12/30 & 40.0\% \\
P7 & MS  & 1--2 yrs     & 25--35 & 17/30 & 56.7\% \\
P8 & MS  & 5+ yrs       & 35--45 & 24/30 & 80.0\% \\
P9 & PhD & 2--5 yrs     & 25--35 & 22/30 & 73.3\% \\
P10 & BS & 1--2 yrs     & 18--25 & 16/30 & 53.3\% \\
P11 & PhD & 5+ yrs      & 45--55 & 26/30 & 86.7\% \\
P12 & BS  & \textless 1 yr & 18--25 & 11/30 & 36.7\% \\
P13 & MS  & 2--5 yrs     & 25--35 & 20/30 & 66.7\% \\
P14 & PhD & 1--2 yrs     & 25--35 & 22/30 & 73.3\% \\
P15 & BS  & \textless 1 yr & 18--25 & 8/30 & 26.7\% \\
P16 & PhD & 5+ yrs       & 35--45 & 25/30 & 83.3\% \\
P17 & MS  & 2--5 yrs     & 25--35 & 23/30 & 76.7\% \\
P18 & PhD & 5+ yrs       & 45--55 & 24/30 & 80.0\% \\
P19 & BS  & \textless 1 yr & 18--25 & 7/30 & 23.3\% \\
P20 & MS  & 1--2 yrs     & 25--35 & 18/30 & 60.0\% \\
\midrule
\multicolumn{5}{l}{\textbf{Expert Average (5+ years experience):}} & \textbf{79.4\%} \\
\multicolumn{5}{l}{\textbf{All Participants Average:}} & \textbf{57.2\%} \\
\bottomrule
\end{tabular}
\end{table}

\begin{table}[htbp]
  \caption{Topic Coverage and Source Papers}
  \label{tab:sources}
  \centering
  \small
  \renewcommand{\arraystretch}{1.25}
  \begin{tabular}{@{}p{2.9cm}p{7.3cm}p{1.4cm}p{1.1cm}@{}}
    \toprule
    \textbf{Topic} & \textbf{References} & \textbf{Years} & \textbf{\#} \\
    \midrule
    Basic Concepts & \cite{aharonov1998quantum, terashima2005nonunitary, arrazola2022universal, williams1998automated, hayward2008quantum, xu2023securing, peterer2015coherence, schuld2019quantum, gudder1983hilbert, biard2021increasing, kowalski2023quantum, del2025features, younis2022quantum, hua2023qasmtrans, sikorskigpu, nielsen2010quantum, preskill2018quantum, deutsch1985quantum, feynman1982simulating, kjaergaard2020superconducting, bharti2022noisy, zurek2003decoherence, deutsch1998quantum, bennett1993teleporting, quinton2025quantum, phillipson2024quantum, cirac1995quantum, benioff1980computer, giovannetti2008quantum, abughanem2025ibm, farhi2000quantum, divincenzo2000physical, karni2026quantum, lloyd1996universal, knill2001scheme, king2025beyond, halimeh2025cold} & 1980--2026 & 37 \\
    \midrule
    Gates \& Circuit Design & \cite{zhang2022escaping, ren2024hardware, peham2022equivalence, kusyk2021survey, ostaszewski2021reinforcement, rosa2025optimizing, meulen2026evaluating, divincenzo1998quantum, kalloor2024quantum, senapati2024pqml, cao2012quantum, yang2026minimization, venturelli2018compiling, barenco1995elementary, vandersypen2001experimental, steane1999efficient, laflamme2002nmr, cory2000nmr, cross2019validating, linke2017experimental, maslov2008quantum, mckay2018qiskit, hashim2021randomized, zulehner2018efficient, wille2019mapping, murali2019noise} & 1995--2026 & 26 \\
    \midrule
    Quantum Machine Learning & \cite{wittek2014quantum, bowles2024better, mandava2026quantum, vishwakarma2024qiskit, ranga2024quantum, rath2024quantum, biswas2025data, bischof2025hybrid, kreplin2024reduction, afane2026differentiable, chinzei2024resource, afane2025atp, yu2024quantum, schuld2015introduction, havlicek2019supervised, cerezo2021variational, biamonte2017quantum, schuld2018supervised, farhi2018classification, dunjko2018machine, benedetti2019parameterized, lloyd2018quantum, beer2020training, huang2021power, rebentrost2014quantum, grant2018hierarchical, cong2019quantum, schuld2020circuit, amin2018quantum, perdomo2018opportunities, moll2018quantum, arrazola2019quantum, romero2017quantum, tacchino2019artificial, abbas2021power} & 2014--2026 & 35 \\
    \midrule
    Distributed Computing & \cite{cuomo2020towards, cacciapuoti2020quantum, wehner2018quantum, kimble2008quantum, diadamo2022distributed, dahlberg2019link, caleffi2022quantum, pompili2021realization, van2016path, munro2015inside, meter2013path, cirac1997quantum, perseguers2013quantum, pant2019routing, ishizaka2008asymmetric, simon2015theoretical, laurat2007entanglement, van2020quantum, lemos2012quantum, pirandola2019advances, takeda2023wavelength, khatri2021principles, bhaskar2020experimental, kozlowski2020designing, muralidharan2016optimal, promponas2025compiler} & 1993--2025 & 35 \\
    \midrule
    Quantum Security & \cite{dhar2024securing, mehic2023quantum, chu2023qdoor, zhao2024bridging, xu2023securing, krawec2024finite, zhang2022experimental, xu2024security, tan2025qubithammer, sahu2024state, ralegankar2021quantum, kalaivani2021enhanced, pirandola2020advances, bernstein2017post, lo2014secure, bennett2014quantum, xu2020secure, ekert1991quantum, bennett1992quantum, scarani2009security, lo1999unconditional, shor2000simple, mayers2001unconditional, renner2008security, wootters1982single, diamanti2016practical, cherkaoui2026categorical} & 1982--2026 & 27 \\
    \midrule
    Error Correction & \cite{fowler2012surface, shor1995scheme, lidar2013quantum, terhal2015quantum, aharonov2008fault, chiaverini2004realization, reed2012realization, bombin2006topological, gottesman1997stabilizer, nielsen2021fault, steane1996error, kitaev1997quantum, preskill1998reliable, bacon2006operator, aliferis2006quantum, calderbank1996good, steane1999efficient, dennis2002topological, acharya2024quantum, barends2014superconducting, kelly2015state, cory1998experimental, divincenzo1996fault, bravyi2005universal, albert2019performance, bennett1996mixed, gambetta2017building, mcewen2023removing, takita2017experimental, wang2026demonstration} & 1995--2026 & 30 \\
    \midrule
    Quantum Algorithms & \cite{montanaro2016quantum, mosca2008quantum, childs2010quantum, hastings2014improving, gheorghiu2025quantum, krovi2023improved, jin2023time, qiang2021implementing, benedetti2021hardware, du2022quantum, motta2020determining, grover1996fast, shor1999polynomial, harrow2009quantum, ambainis2010quantum, kitaev1995quantum, nayak1999quantum, childs2003exponential, cleve1998quantum, farhi2000quantum, nielsen1998quantum, aharonov2003adiabatic, bennett1997strengths, deutsch1992rapid, nielsen2002quantum, brassard2002quantum, jordan2009fast, reichardt2009span, wiebe2012quantum, aaronson2011computational, an2026fast} & 1995--2025 & 31 \\
    \bottomrule
  \end{tabular}
\end{table}

\end{document}